%% file: main.tex
\documentclass[11pt]{article}
\usepackage[preprint]{acl}
\usepackage{times}
\usepackage{latexsym}
\usepackage[T1]{fontenc}
\usepackage[utf8]{inputenc}
\usepackage{microtype}
\usepackage{inconsolata}

\usepackage{graphicx} 
\usepackage{booktabs}
\usepackage{multirow}
\usepackage{pgfplots}
\pgfplotsset{compat=1.18}
\usepackage{subcaption}
\usepackage{amsmath}

\title{CALRK-Bench: Evaluating Context-Aware\\
Legal Reasoning in Korean Law}
\author{
JiHyeok Jung$^{1}$ \quad
Taeyoung Yoon$^{2}$ \quad
Hyunsouk Cho$^{3}$ \\
\texttt{ji9759@kaist.ac.kr} \quad
\texttt{yty@ajou.ac.kr} \quad
\texttt{hyunsouk@ajou.ac.kr} \\
$^{1}$KAIST AI \\
$^{2}$Law School, Ajou University \\
$^{3}$Department of Artificial Intelligence, Ajou University
}

\begin{document}
\input{macros}

\maketitle
\input{Sections/0_abs}

\section{Introduction}
\input{Sections/1_intro}

\section{Related Work}
\input{Sections/2_rel}

\section{Context-Aware Legal Reasoning Benchmark}
\input{Sections/3_method}

\section{Experiments}
\input{Sections/4_exp}

\section{Conclusion}
\input{Sections/5_con}

\section*{Limitations}
\input{Sections/6_limitation}

\bibliography{ref}

\clearpage
\appendix
\section{Appendix}
\input{Sections/7_appendix}
\input{Sections/8_appendix_example}
\end{document}

%% file: macros.tex
\newcommand{\eg}{e.g.}
\newcommand{\ie}{i.e.}

\newcommand{\ours}{\textbf{CALRK-Bench}}

\newcommand{\ttime}{\textbf{Type-TCR}}
\newcommand{\tsuff}{\textbf{Type-ISR}}
\newcommand{\tshift}{\textbf{Type-JSA}}

\newcommand{\todoc}[2]{{\textcolor{#1}{\textbf{#2}}}}
\newcommand{\todoorange}[1]{\todoc{orange}{\textbf{[[#1]]}}}
\newcommand{\hist}[1]{\todoorange{hist: #1}}

\newcommand{\romannum}[1]{\romannumeral #1}
\newcommand{\com}{\textcolor{red}}
\newcommand{\revise}{\textcolor{blue}}

%% file: Sections/0_abs.tex
\begin{abstract} Legal reasoning requires not only the application of legal rules but also an understanding of the context in which those rules operate. However, existing legal benchmarks primarily evaluate rule application under the assumption of fixed norms, and thus fail to capture situations where legal judgments shift or where multiple norms interact. In this work, we propose CALRK-Bench, a context-aware legal reasoning benchmark based on the legal system in Korean. CALRK-Bench evaluates whether models can identify the temporal validity of legal norms, determine whether sufficient legal information is available for a given case, and understand the reasons behind shifts in legal judgments. The dataset is constructed from legal precedents and legal consultation records, and is validated by legal experts.
Experimental results show that even recent large language models consistently exhibit low performance on these three tasks. CALRK-Bench provides a new stress test for evaluating context-aware legal reasoning rather than simple memorization of legal knowledge. Our code is available at ~\url{https://github.com/jhCOR/CALRKBench}. \end{abstract}

%% file: Sections/1_intro.tex
Legal reasoning is the process of deriving judicial conclusions by interpreting and applying legal norms based on established facts~\cite{fan2025lexam}. With the advancement of Large Language Models (LLMs), such reasoning structures have increasingly been utilized in various legal tasks such as legal judgment prediction, statutory interpretation, and case analysis~\cite{akarajaradwong2025nitibench,fan2025lexam,aletras2016predicting, niklaus2021swiss}. LLMs demonstrate a certain level of competence in rule-application-oriented tasks when both the factual circumstances and the applicable legal provisions are explicitly provided or assumed to be fixed~\cite{achiam2023gpt}.

\begin{figure}[!t]
    \centering
    \includegraphics[width=\columnwidth]{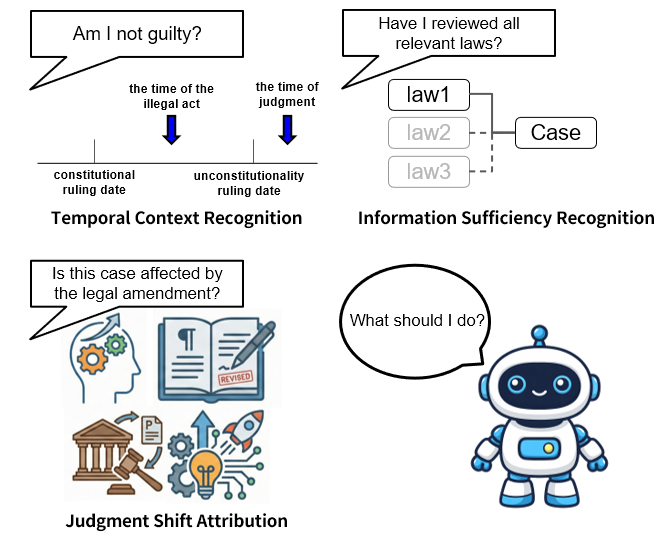}
    \caption{Overview of the Benchmark.}
    \label{fig:pipeline}
\end{figure}

Many legal benchmarks have already evaluated tasks such as statutory interpretation, legal question answering, and judgment prediction. Although they take different forms on the surface, most studies formalize legal reasoning as the process of applying a given rule to given facts, which reflects the IRAC (Issue–Rule–Application–Conclusion) structure well~\cite{metzler2002importance}. In particular, previous studies have mainly focused on knowledge recall ability~\cite{fei2024lawbench, li2024lexeval} or the reproduction of judgment outcomes~\cite{hwang2022multi}. 

However, such studies cannot reflect situations in which laws are amended or repealed, or situations in which the governing norm remains the same but judicial decisions change. These changes are typically treated as contextual factors because they are not directly observable from the immediate facts of a case. These concepts are also well explained in Hart’s classical legal theory~\cite{hart2012concept}. Hart defined law not as a mere collection of rules governing conduct, but as a combination of primary rules governing conduct and secondary rules governing the validity and change of rules. In particular, the rule of recognition and the rule of change determine which norms constitute valid law and when legal norms are created or modified. These systemic characteristics show that legal judgment is not merely the application of rules but a contextual determination that considers which norms are valid at a given time and how their legal effects interact.

To address this gap, we propose \textbf{CALRK-Bench (Context-Aware Legal Reasoning in Korean Law)}, a benchmark that conceptualizes legal judgment as a multi-stage process prior to rule application and evaluates whether models can (i) identify legally meaningful changes (\tshift), (ii) reason about temporal validity and scope of application (\ttime), and (iii) assess the sufficiency of legal information (\tsuff) prior to judgment. However, these capabilities have rarely been evaluated in IRAC-centered benchmarks.

To enable reliable evaluation of these capabilities, the benchmark is constructed based on the Korean legal system, which provides practical advantages for such analysis. As a civil law system, Korean legal norms are primarily articulated in written statutes, making the temporal validity and applicability of norms relatively clear and amenable to rule-based modeling. By contrast, common-law systems rely more heavily on precedent-based doctrinal development, where legal change often emerges gradually through judicial interpretation, making doctrinal shifts harder to formalize into precise evaluation criteria.

CALRK consists of 100, 279, and 100 stress-test questions for \ttime, \tsuff, and \tshift, respectively. Considering the inherent rarity of clearly classifiable doctrinal change cases, the benchmark prioritizes rigor over scale and focuses on collecting high-quality instances grounded in real legal sources. The dataset is constructed from legal precedents and consultation records, with carefully designed problem instances that reflect realistic legal reasoning scenarios. In particular, for information sufficiency tasks, we employ a hard negative mining procedure to construct legally plausible but misleading distractors based on co-citation patterns in precedents and conflicting legal conclusions. Such co-cited provisions reflect real legal scenarios in which competing arguments rely on different statutory grounds for the same case, requiring the model to determine which legal basis is actually governing and whether the provided information is sufficient for reaching a judgment. The resulting dataset, while emphasizing quality over scale, is comparable in size to prior legal benchmarks such as GoldCoin~\cite{fan2024goldcoin} and the NitiBench-Tax subset of NitiBench~\cite{akarajaradwong2025nitibench}.

Experimental results show that even state-of-the-art LLMs exhibit systematic failures across all three tasks. Models struggle to reason about the temporal validity of norms, fail to recognize when available information is insufficient, and show biases in attributing the causes of judgment shifts. These findings highlight fundamental limitations of current LLMs in context-aware legal reasoning, which CALRK-Bench is designed to evaluate.

%% file: Sections/2_rel.tex
\subsection{General Legal Benchmarks}

The expansion of Large Language Models (LLMs) into the legal domain has produced numerous evaluation benchmarks and datasets. Representative efforts such as MMLU, LexGLUE, LegalBench, and LEXAM~\cite{chalkidis2022lexglue, pipitone2024legalbench, hendrycks2020measuring, fan2025lexam} cover tasks including statute interpretation, legal QA, document classification, and judgment prediction, establishing the foundation for systematic legal reasoning evaluation.

Despite differences in scope, most assume a fixed and predetermined governing rule rooted in the IRAC (Issue--Rule--Application--Conclusion) framework. Evaluation, therefore, measures rule application accuracy rather than whether the normative framework itself remains valid or applicable.

\subsection{Temporal Context Benchmarks}

Several studies have examined how large language models handle the temporal aspects of legal information. 
LawShift~\cite{han2025lawshift} shows that LLMs often fail to properly adjust their judgments when legal norms change due to statutory revisions. 
Other studies address temporality more indirectly. 
For example, LexTime~\cite{barale2025lextime} analyzes the temporal ordering of events in legal texts, while MultiEURLEX~\cite{chalkidis2021multieurlex} highlights temporal concept drift in the distribution of legal labels.

\subsection{Sufficient Context Benchmarks}
In real-world legal reasoning, various legal and factual information may be involved, and in such situations, it may be necessary to recognize that additional information is required before reaching a conclusion. Some studies have addressed this issue indirectly. For example, previous work analyzes split-vote decisions of the European Court of Human Rights (ECHR) and measures disagreement among judges and case difficulty, evaluating how well models recognize such uncertainty~\cite{xu2024through}. However, these studies focus on uncertainty awareness or calibration rather than directly evaluating whether the provided legal information is sufficient for making a judgment.

\subsection{Shift Context Benchmarks}

Recent studies have also explored disagreement and change in legal judgments. 
For example, previous work investigates whether LLMs can identify overruling relationships between judicial precedents in long legal documents~\cite{zhang2025llms}. 
The results show that even when the overruling is explicitly stated, models often rely on shallow heuristics and struggle to correctly identify such relationships.

%% file: Sections/3_method.tex
In this section, we formalize the problem and introduce our benchmark for evaluating context-aware legal reasoning in LLMs.

\subsection{Task Formulation}

We assess whether legal LLMs can recognize contextual factors in legal reasoning through three tasks.

\textbf{\ttime: Temporal Context Recognition}

Given a judgment summary derived from real legal precedents and a description of a subsequent normative change, the model must select the point in time at which a conclusion similar to the given case would be possible. The dataset is constructed by collecting cases involving clear legal changes, including constitutional invalidation or constitutional nonconformity leading to the nullification of statutes, statutory amendments, and changes in judicial precedents.

Each instance is based on either a case governed by the pre-change norm or one governed by the post-change norm, paired with an explicit description of the legal change. To construct answer choices, we define two reference time points ($A_1$, $A_2$) that characterize each type of legal change (e.g., decision dates in constitutional rulings, promulgation and enforcement dates in statutory amendments, and precedent change decisions). These reference points partition the timeline into three intervals.

Answer choices are generated by enumerating combinations of the time of the illegal act and the time of trial across these intervals. Depending on their positions relative to $A_1$ and $A_2$, each combination determines whether the pre-change or post-change norm applies. The correct option corresponds to the temporal configuration that reproduces the original case outcome, while the remaining options reflect alternative configurations governed by a different norm.

This task evaluates whether models can reason about the temporal validity of legal norms, rather than assuming that norms are fixed across time.

\textbf{\tsuff: Information Sufficiency Recognition}

The \tsuff\ task evaluates the ability to determine whether the provided legal provisions are sufficient to answer the given legal consultation question. For this purpose, three conditions are defined according to the composition of the provisions provided.

\begin{itemize}
\item Sufficient: A condition in which all key legal provisions necessary to answer the question are provided
\item Partial: A condition in which only some of the legal provisions necessary to answer the question are provided
\item Insufficient: A condition in which the legal provisions necessary to answer the question are not provided, and instead provisions that may appear together in an actual legal context but do not constitute a direct basis for the question are provided
\end{itemize}

This task evaluates whether the model can identify the scope of legal information required by the given question and determine whether the provided provisions alone are sufficient.

In the Type-ISR task, the model is required to select the correct answer from four options. One of these options is always fixed to indicate that ``no additional legal reference is needed.'' For questions requiring additional legal provisions, one of the statutes included in the expert answer is provided as the correct option. Furthermore, all samples are constructed by excluding precedent-dependent cases, ensuring that each consultation can be answered solely based on statutory provisions.

The remaining incorrect options are constructed through a hard negative mining procedure. These distractors are designed to (1) have a high likelihood of co-occurring with the correct statute in real legal contexts, and (2) be legally plausible alternatives that can confuse the model, rather than being trivially irrelevant. In particular, statutes that exhibit conflicting legal effects or restrict each other's scope of application are well-suited for this purpose. Such statutes are often observed in real judicial cases where opposing legal arguments are made under the same factual circumstances. Accordingly, candidate distractors are derived from statutes co-cited in relevant precedents.

Concretely, given a gold statute extracted from expert answers, we collect candidate statutes that co-occur with the gold statute in precedents where it is cited. We then utilize an LLM to filter samples for which the question can be correctly answered using only the gold statute. For each candidate statute, we compare the conclusion derived from the gold statute with that derived from the candidate statute, and retain only those candidates that induce conflicting conclusions as hard negatives.

Through this process, the constructed distractors are not merely irrelevant alternatives but legally plausible yet incorrect statutes that can actively mislead the model. Further implementation details and configurations are provided in the Appendix.

\textbf{\tshift: Judgment Shift Attribution}
Legal judgments do not always converge to the same conclusion; depending on changes in social and institutional contexts, outcomes may differ even for the same legal issue. Such variations arise from factors including amendments to legal norms, changes in precedents, and shifts in social or factual conditions. In practice, determining whether a case falls under such change factors is often a key issue.

We categorize potential causes into four types, inspired by prior discussions of legal change in the literature: shifts in normative foundations, modifications of legal texts, reinterpretation through precedent, and changes in factual or socio-economic conditions~\cite{hart2012concept,maccormick2016interpreting,schauer1987precedent}.

These categories are designed to be mutually exclusive. Although precedent changes may be influenced by broader social shifts, they function as binding interpretive rules within the legal system. In contrast, social or technological changes may affect judicial reasoning but do not themselves create legally binding norms.

The causes of judgment change are defined as follows:
\begin{itemize}
\item \textbf{Normative premises}: changes in general legal consciousness of society, conventions, and social or normative foundations
\item \textbf{Normative text}: enactment or amendment of relevant legal provisions, including addition/deletion, and loss of legal validity due to unconstitutionality or constitutional nonconformity
\item \textbf{Interpretive rules}: explicit changes in judicial interpretation through precedent reversal, typically by Supreme Court en banc decisions, which establish new interpretive rules that subsequent courts are expected to follow.
\item \textbf{Factual premises}: the norm and interpretive rules themselves remain the same, but substantive values or technological/industrial environments are created or changed
\end{itemize}

Given a judgment summary and a description of change, the model must identify the direct cause of the judgment change. This task evaluates whether models can distinguish legally distinct sources of change, rather than relying on vague or generalized explanations. Through this, the benchmark evaluates legal context understanding ability that goes beyond simple prediction of judgment outcomes and identifies the structural causes of judgment change.

\paragraph{Expert Validation}

This dataset was constructed with the important goal of securing a legally rigorous composition. In the process of constructing options including temporal contexts, when the conclusion was not clear from legal provisions alone, we confirmed whether cases actually examined under the relevant temporal conditions existed in Supreme Court precedents and conducted a review by a law professor. In addition, during the case collection process, a researcher manually selected seed cases by reviewing commentaries provided in legal databases and referring to related legal provisions and case law studies.

Finally, the task definitions and option construction were reviewed by a law professor and further validated through consultation with a legal practitioner. During the process of defining task categories and designing options, repeated feedback was reflected to revise the task definitions for legal ambiguities or potential overlaps between category boundaries that were identified. In particular, an intensive review was conducted regarding the terminology of prerequisite conditions such as the “date of constitutional decision” presented for judging temporal normative application in \ttime\ and the category system for classifying the causes of judgment change in \tshift.

Through this repeated review process, the task definitions and option system were refined to a level that is legally consistent and acceptable to experts.

\subsection{Dataset Statistics}

The overall scale of this benchmark may appear small compared to typical NLP datasets, reflecting characteristics of the legal domain. In an actual legal system, explicit changes to existing precedents or socially significant statutory amendments occur relatively rarely. For example, en banc decisions of the Korean Supreme Court amounted to only 498 cases over the 30-year period from 1993 to 2023~\cite{ART003230177}. Precedents in which such changes can be clearly observed are even more limited.

In addition, to accurately evaluate the reasoning ability of models, this benchmark intentionally excluded cases where legal interpretation is controversial or where only minor changes exist. This is to secure the rigor of the evaluation data by selecting only cases where supporting materials such as commentaries and case law studies are abundant and where distinctions can be clearly made, rather than prioritizing expansion of data scale.

During the data construction process, precedents from Lawnb\footnote{\url{https://www.lawnb.com}}, Korea Legal Aid Corporation consultation data\footnote{\url{https://www.klac.or.kr}}, precedents provided by AI Hub\footnote{\url{https://aihub.or.kr}}, and judicial precedents from the Korean Law Information Center\footnote{\url{https://www.law.go.kr}} operated by the Ministry of Government Legislation were utilized. The data were used in accordance with the respective licenses and terms of use. To comply with these restrictions, the original text of the precedents was not disclosed, and consultation questions were used in their original form.

For the case data, 52 precedents before the normative change and 48 precedents after the change were collected. Case pairs were constructed based on possible precedent combinations. For the \tshift\ task, 33 precedents following the norms before the change and 38 precedents following the norms after the change were used. Case pairs were constructed from combinations of these precedents, and pairs were selected so that the distribution across categories was as uniform as possible. As a result, 28 items for Normative premises, 28 items for Normative text, 26 items for Interpretive rules, and 18 items for Factual premises were constructed. For the \ttime\ task, only cases involving constitutional invalidation, statutory amendment, and precedent change were further filtered, resulting in 29 precedents from before the change and 25 precedents from after the change being used. Finally, 44, 8, and 48 questions were constructed for constitutional invalidation, statutory amendment, and precedent change case, respectively.

For the \tsuff\ task, a total of 279 questions were generated based on 22 consultation cases. 
Incorrect answer options were constructed through a Hard Negative Mining procedure using legal provisions that co-occur in precedents with the reference statute serving as the correct answer. 
Among these, only provisions that are not required to answer the question were retained. 
Such provisions may appear together in an actual legal context but do not constitute a direct legal basis for the question. Detailed procedures of the data construction process are provided in the Appendix.

%% file: Sections/4_exp.tex
\subsection{Experimental Settings}

\paragraph{Base dataset}

Each item in CALRK-Bench is composed as a multiple-choice question that selects one among four options. Therefore, the random choice baseline accuracy is 25\%. Evaluation was conducted based on overall accuracy, and together with overall accuracy, the accuracy for each type—temporal context, sufficiency context, and change context—was also reported.

\input{Sections/tab_main}

In addition, some models such as GPT and Gemini can adjust reasoning effort, and this setting controls the depth of reasoning. In this study, in order to analyze the effect of reasoning effort on performance, multiple reasoning levels were evaluated together for models where this setting is provided.

\paragraph{Models and evaluation metric}

The following LLM models were used for evaluation: 
gpt-5-2025-08-07, gemini-3-flash-preview, llama-3.3-70b-instruct, 
qwen3-30b-a3b-thinking-2507, qwen3-30b-a3b-instruct-2507.

For open-source models, we use deterministic decoding (temperature=0, top-p=1) and report results from a single run with a fixed seed. For API-based models, we report average results over three runs due to stochasticity.

\subsection{Overall Performance}
Even state-of-the-art LLMs struggle to solve CALRK-Bench. 
Table~\ref{tab:overall_results} shows the overall accuracy and type-wise performance of the evaluated models. Looking at type-wise performance, different failure patterns are observed in each task. 

In the \ttime\ task, models struggle to reason over temporal context that requires jointly considering the timing of normative changes, the time of the illegal act, and the time of trial. In particular, the task requires identifying the time interval in which the same conclusion as the given case would hold under the applicable norm, but models often fail to correctly integrate these temporal factors. This difficulty is especially pronounced in cases where the pre-change norm should be applied.

In the \tsuff\ task, when looking only at overall accuracy the performance appears relatively high, but detailed analysis results show that models show high accuracy under the \textit{Sufficient condition} where additional legal reference is not required, whereas performance greatly decreases under the \textit{Partial condition} or the \textit{Insufficient condition}. In addition, it was generally observed that performance tends to decrease as reasoning effort increases.

In the \tshift\ task, models exhibit systematic category-wise biases, often over-attributing judgment changes to particular categories instead of accurately distinguishing among different causes.

These results suggest that current LLMs may have learned a considerable portion of the content of legal norms themselves, but may still show limited performance in contextual legal reasoning abilities such as identifying applicable norms by considering the temporal context of events or evaluating whether the given legal information is sufficient for judgment.

In order to analyze these failure patterns observed in each task in more detail, the next section conducts additional analyses and validation experiments by type.

\subsection{Type-wise Results}

\begin{table}[t]
\centering
\begin{tabular}{lccc}
\toprule
Model & before & after & avg \\
\midrule
GPT-5 (minimal) & 0.053 & 0.600 & 0.327 \\
GPT-5 (low) & 0.427 & 0.747 & 0.587 \\
GPT-5 (medium) & 0.500 & 0.747 & 0.623 \\
GPT-5 (high) & 0.493 & 0.767 & 0.630 \\
\bottomrule
\end{tabular}
\caption{\ttime\ performance under different reasoning effort levels of GPT-5.}
\label{tab:gpt5_reasoning}
\end{table}

\paragraph{\ttime: Temporal Context Recognition}

Models exhibit a large performance gap between cases governed by the pre-change norm and those governed by the post-change norm. 
Table~\ref{tab:overall_results} reports model accuracy under these two conditions.

Across all models, accuracy is substantially higher when the post-change norm applies than when the pre-change norm must be considered. In addition, overall performance in the temporal context recognition task tends to improve as reasoning effort increases. These results suggest that temporal context reasoning requires not simple rule memorization, but precise reasoning about the relationship between the time of the act, the time of the trial, and the timing of normative change. 

In particular, in cases where the norm after change is applied, it is a relatively simple problem because the norm after legal amendment or precedent change can be applied. In contrast, in cases where the norm before change is applied, there are cases where the relationship between the time of the act and the time of the trial must be considered simultaneously, and in such cases a tendency was observed where model performance greatly decreases for models with weaker reasoning capabilities.

For example, even if the trial takes place after a constitutional decision, if the time of the act corresponds to before the constitutional decision, a different conclusion such as dismissal of prosecution rather than a guilty or not-guilty judgment is derived.

These results show that the difficulty of the \ttime\ task lies not in simple rule application or memorization, but in a complex reasoning problem that requires jointly considering the time of the unlawful act, the time of the trial, and the type of normative change (e.g., interpretive change, constitutional invalidation).

\begin{table}[t]
\centering
\begin{tabular}{lcc}
\toprule
Model & Acc. & $\Delta$ \\
\midrule
GPT-5 (m) & \textbf{0.717} & -- \\
GPT-5 (l) & 0.651 & -0.066 \\
\midrule
LLaMA 3.3 70B & \textbf{0.682} & -- \\
LLaMA 3.3 70B + CoT & 0.664 & -0.018 \\
\midrule
Qwen3-30B-A3B & \textbf{0.757} & -- \\
Qwen3-30B-A3B + CoT & 0.720 & -0.037 \\
Qwen3-30B-A3B (T) & 0.654 & -0.103 \\
\bottomrule
\end{tabular}
\caption{
Performance under the \textit{Partial} condition in \tsuff. 
Acc denotes accuracy under the \textit{Partial} condition. 
$\Delta$ indicates the change in accuracy between the baseline model and its reasoning-augmented variant (e.g., CoT prompting or thinking-mode model).
}
\label{tab:need_partial}
\end{table}

\paragraph{\tsuff: Information Sufficiency Recognition}
Models generally achieve high accuracy when all necessary legal provisions are provided, but their performance drops substantially when only partial information is available. To further analyze this phenomenon, we report model performance under the \textit{Partial} condition in Table~\ref{tab:need_partial}. For several models, we additionally evaluated a Chain-of-Thought (CoT) prompting variant in which the model was instructed to reason step by step and present the final answer on the last line in the format ``Answer: (final answer)''. The instruction was provided in Korean. In particular, under the \textit{Partial} condition, accuracy tends to decrease as the reasoning effort setting increases. This suggests the possibility that the model fails to recognize that additional information is required in situations where information is insufficient, and instead falls into early conclusions or excessive confidence by relying on internal reasoning based on partially provided knowledge.

\paragraph{Context Injection Experiment}

The performance drop observed under the Partial condition when using reasoning and Chain-of-Thought suggests that the model may have reached incorrect conclusions by relying on parametric knowledge even when the provided information is insufficient. Prior work shows that parametric knowledge may dominate when the input context contains related cues~\cite{kortukov2024studying}, while the model’s trust in external evidence depends on the coherence and convincingness of the context~\cite{xie2023adaptive}. In our experiments as well, the performance under the \textit{Partial} condition was generally lower than that under the Insufficient condition, which may be because the partially provided information in the \textit{Partial} condition contains keywords similar to the parametric knowledge required to solve the task.

Meanwhile, prior studies have reported that when explicit contextual knowledge is provided as input, the use of parametric knowledge may be suppressed~\cite{cheng2024understanding}. Motivated by this observation, this study conducted a variant experiment in which an additional context that does not provide direct clues to the answer was injected. Specifically, Chinese legal QA sentences not directly related to Korean legal problems were randomly selected from the LexEval benchmark~\cite{li2024lexeval} and inserted before the input prompt. In addition, to control for the simple effect of increased context length and to examine the influence of context coherence on the results, we also conducted a comparison experiment by inserting random strings.

\begin{figure}[t]
    \centering
    \includegraphics[width=\columnwidth]{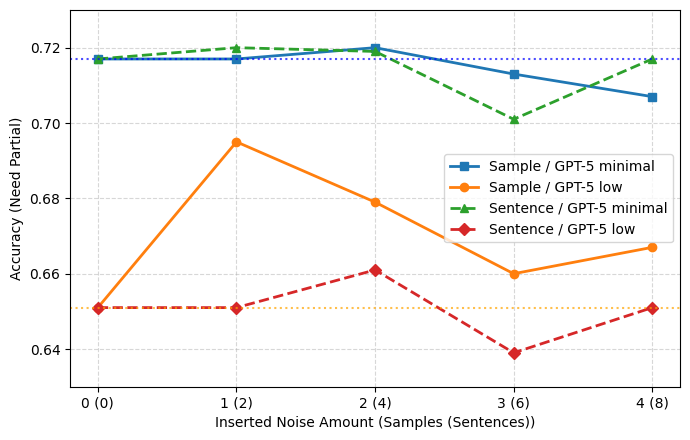}
    \caption{For the experiments involving context addition, the LexEval data were randomly sampled or the strings were randomly generated. Results averaged over three runs.}
    \label{fig:type2_analysis}
\end{figure}

Under the Partial condition, GPT-5 with the low reasoning effort setting showed an accuracy improvement of up to 4.4\%, as shown in Figure~\ref{fig:type2_analysis}. In contrast, no meaningful change was observed under the minimal reasoning setting, and inserting random strings resulted in almost no performance difference. These results suggest that even contexts that do not directly contribute to solving the task can still influence model behavior under certain settings, indicating that some errors may stem from reliance on internal parametric knowledge (PK).

\begin{figure}[t]
\centering
\begin{subfigure}{0.48\linewidth}
    \centering
    \includegraphics[width=\linewidth]{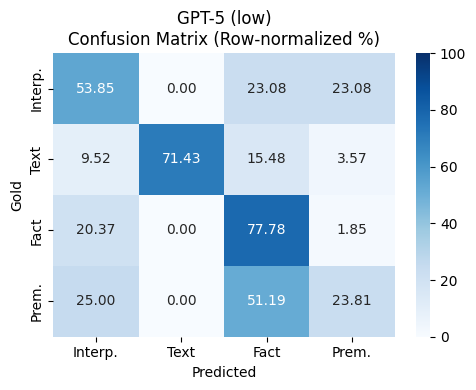}
    \caption{GPT}
\end{subfigure}
\hfill
\begin{subfigure}{0.48\linewidth}
    \centering
    \includegraphics[width=\linewidth]{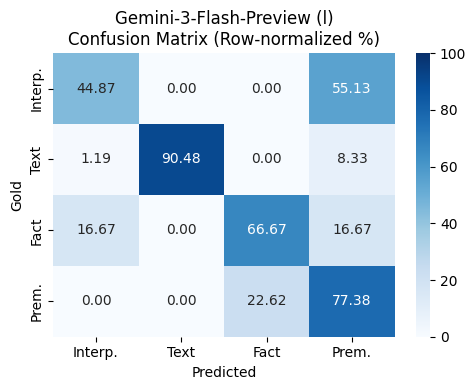}
    \caption{Gemini}
\end{subfigure}

\caption{Row-normalized confusion matrices for \tshift. Each value represents the percentage of predictions within the same gold label.}
\label{fig:type3_confusion}
\end{figure}

\paragraph{\tshift: Judgment Shift Attribution}

Experimental results showed that in most models a tendency was observed where specific options are consistently selected more frequently, as illustrated in Figure~\ref{fig:type3_confusion}. For example, Gemini showed the highest overall accuracy, but it tended to select \textit{Normative premises} at a relatively high rate, while GPT models showed a pattern of selecting \textit{Factual premises} more frequently.

These results suggest the possibility that, rather than distinguishing the structural cause of judgment change through the specific context of the case, the models may exhibit a heuristic preference for certain types of change. One possible explanation may be related to the bias of legal systems present in the training data. In practice, the way legal judgment changes occur differs depending on the legal system. For example, in civil law systems such as Korea, changes in judicial outcomes often occur through institutional procedures such as explicit statutory amendments or precedent reversals by the Supreme Court en banc. In contrast, in common law systems such as those of the United Kingdom or the United States, social changes or shifts in normative perceptions are often reflected more directly through judicial interpretation of precedents.

The training data of large language models contain a high proportion of English-language materials, and such a data composition may make the models relatively more familiar with patterns of change observed in common law systems. As a result, there is a possibility that the models tend to more naturally select categories such as social change or shifts in normative perceptions.

%% file: Sections/tab_main.tex
\begin{table*}[t]
\centering
\setlength{\tabcolsep}{2.5pt}

\begin{tabular}{lccc|cccc|ccccc}
\toprule
 & \multicolumn{3}{c}{\ttime} & \multicolumn{4}{c}{\tsuff} & \multicolumn{5}{c}{\tshift} \\
\cmidrule(lr){2-4} \cmidrule(lr){5-8} \cmidrule(lr){9-13}
Model
& Pre. & Post. & Avg
& Suff. & Insuff. & Part. & Avg
& Prem. & Text & Interp. & Fact & Avg \\
\midrule

GPT-5* (m)
& 0.053 & 0.600 & 0.327
& 0.885 & \underline{0.753} & \textbf{0.717} & \textbf{0.762}
& 0.226 & 0.679 & 0.244 & \textbf{0.944} & 0.487 \\

GPT-5* (l)
& 0.427 & \underline{0.747} & \underline{0.587}
& 0.973 & \underline{0.756} & 0.651 & \underline{0.754}
& 0.238 & 0.714 & \textbf{0.539} & 0.778 & 0.547 \\

Gemini Flash* (m)
& 0.200 & 0.727 & 0.463
& \textbf{1.000} & 0.721 & 0.636 & 0.737
& \underline{0.702} & \underline{0.821} & 0.154 & \textbf{0.944} & \underline{0.637} \\

Gemini Flash* (l)
& \textbf{0.720} & \textbf{0.820} & \textbf{0.770}
& \underline{0.993} & 0.737 & 0.648 & 0.748
& \textbf{0.774} & \textbf{0.905} & 0.449 & 0.667 & \textbf{0.707} \\

LLaMA 3.3 70B
& 0.000 & 0.620 & 0.310
& 0.633 & 0.707 & 0.682 & 0.685
& 0.536 & 0.107 & 0.077 & \underline{0.833} & 0.350 \\

Qwen3-30B-A3B T
& \underline{0.480} & 0.580 & 0.530
& 0.878 & 0.715 & 0.654 & 0.720
& 0.107 & 0.179 & \underline{0.462} & 0.389 & 0.270 \\

Qwen3-30B-A3B I
& 0.060 & 0.560 & 0.310
& 0.327 & 0.748 & \textbf{0.757} & 0.677
& 0.288 & 0.250 & 0.000 & 0.556 & 0.250 \\

\bottomrule
\end{tabular}

\caption{
Experiment conducted via OpenRouter. Detailed performance breakdown on CALRK-Bench.
The best score in each column is shown in \textbf{bold}, and the second-best score is \underline{underlined}.
For GPT and Gemini models, we ran the evaluation three times with the seed fixed and report the average results due to internal stochasticity of the models. For open-source models, we used a fixed seed and report results from a single deterministic run.
m: reasoning effort minimal, l: reasoning effort low, T: Thinking model, I: Instruct model. For \tshift, the options were randomly shuffled during evaluation.
}
\label{tab:overall_results}
\end{table*}

%% file: Sections/5_con.tex
In this work, we propose CALRK-Bench, a benchmark designed to evaluate context-aware legal reasoning abilities. Unlike existing legal benchmarks that evaluate rule application under the assumption of fixed norms, CALRK-Bench is designed to evaluate context-aware reasoning abilities required before rule application. Experimental results show that even recent large language models consistently struggle with these tasks, revealing structural limitations in recognizing normative changes and reasoning under contextual legal conditions.

Although this benchmark is constructed based on the Korean legal system, this choice reflects the practical constraint that expert validation is essential in the process of building legal datasets. Situations in which norms change over time and multiple laws and contextual factors must be considered for a single case commonly appear across different legal systems. By leveraging this characteristic, CALRK-Bench evaluates the ability to understand normative changes occurring in legal contexts and reason about them, rather than simple memorization of legal knowledge.

%% file: Sections/6_limitation.tex
This study has several limitations.

First, this benchmark is constructed based on the Korean legal system. Therefore, while the concepts evaluated and the dataset construction methodology may be applicable to other legal systems, directly applying Korean legal data to evaluate legal-domain models specialized for other jurisdictions may be difficult.

Second, cases involving clear normative changes or precedent shifts are relatively rare in real legal systems, which limits the overall scale of the dataset.  In addition, each case was manually verified through expert review to ensure the correctness of legal interpretation and labeling, which further constrains the dataset scale.

Third, to ensure clarity in evaluation, the dataset is primarily constructed from cases where legal changes are relatively explicit. As a result, ambiguous or gradual changes that may arise in real legal judgments are not sufficiently reflected.

In actual legal disputes, even for the same issue, various interpretations and debates may accumulate over a long period before a precedent is established, or existing precedents may be overturned through en banc decisions of the Supreme Court.

In this study, such ambiguous cases were intentionally excluded to ensure consistency and objectivity in evaluation.

Future work could expand the dataset across different legal systems and incorporate cases with more limited supporting evidence by involving multiple legal experts in the labeling process, thereby constructing a dataset that captures a broader range of contextual legal changes.

%% file: Sections/7_appendix.tex
\subsection{Dataset Construction Details}
This section describes the detailed procedure used to construct the CALRK-Bench dataset. In particular, it includes candidate case filtering using LLMs, case summarization, and hard negative mining procedures. All LLM outputs were manually reviewed and revised by the researchers before being included in the final dataset.

\subsubsection{\ttime\ and \tshift\ Data Collection}

For the \ttime\ task and the \tshift\ task, the dataset was constructed by expanding cases centered on seed case pairs. Each seed pair consists of two precedents that represent different legal conclusions before and after a normative change. Based on each seed pair, additional cases sharing the same legal issue were collected to construct a candidate case set.

The scale of similar precedents for the same keyword (case issue) was expanded through the following stepwise procedure.

\paragraph{Keyword-based case collection.}
Using keywords representing the core legal issue of the case, precedents were searched from the AI Hub precedent dataset, resulting in an initial candidate set of 569 cases.

\paragraph{Core-issue identification.}
We used \texttt{gpt-5-2025-08-07} to determine whether the keyword corresponded to the core issue of each case. Cases in which the keyword was mentioned only incidentally were removed, leaving 293 cases.

\paragraph{Case ordering and conclusion similarity judgment.}
Since doctrinal change typically exhibits a temporal pattern, the collected cases were first sorted chronologically. We then used \texttt{gpt-5-2025-08-07} to sequentially determine whether the conclusion of each candidate case was more similar to the norm before the change or the norm after the change in the seed pair, thereby detecting the point at which the conclusion changed.

The model was also required to report a confidence score for the similarity judgment. Cases with confidence lower than 0.6 were removed. This threshold was introduced to efficiently eliminate cases that were clearly unrelated before manual review by researchers. The final determination was made through direct manual inspection by the researchers. Through this process, 136 cases remained.

Finally, a researcher manually reviewed the cases and finalized 100 precedents, including the 20 seed cases.

\subsubsection{\tsuff\ Data Collection and Hard Negative Mining}

The \tsuff\ task evaluates the ability to determine whether the provided legal information alone is sufficient to derive a conclusion. During dataset construction, we used real legal consultation data from the Korea Legal Aid Corporation.

\paragraph{Consultation data collection and filtering.}

A total of 205 question--answer pairs were collected using the keywords ``possible'' and ``permitted''. Among them, all cases that directly referenced judicial precedents in their answers were removed. This is because when a consultation answer is based on a precedent, the legal provisions cited in that precedent may also become necessary information for answering the question.

In contrast, consultation cases that can be answered solely based on statutory provisions typically correspond to situations where a specific law explicitly permits or prohibits an action, making the relevant provisions relatively clear. We further removed cases containing keywords such as ``judgment,'' ``decision,'' ``Constitutional Court,'' leaving 118 question--answer pairs.

Through this filtering process, we obtained the list of legal provisions actually cited by experts in the questions and answers, which served as the gold statutes.

\paragraph{Hard Negative Mining}

Based on the gold statutes extracted from expert answers, we performed hard negative mining to collect candidate incorrect statutes that could compete with the correct statute.

We searched for precedents based on the gold statutes and collected other statutes that co-occurred with the gold statute in those precedents. These co-citation statutes are more likely to appear together with the correct statute in real legal contexts, and therefore have much higher surface-level relevance than randomly selected statutes. This approach allowed us to collect candidate statutes that may appear related but do not directly answer the question or may conflict with the correct legal reasoning.

The variables used in the filtering process are defined as follows. 

\begin{itemize}
\item \textbf{$A$ (gold control)}: conclusion derived from the question + gold statute + expert answer
\item \textbf{$B$ (statute-based gold)}: conclusion derived from the question + gold statute
\item \textbf{$B_{\text{conf}}$ (confounding conclusion)}: conclusion derived from the question + candidate incorrect statute
\end{itemize}

Each variable represents the conclusion generated by the gpt-5-2025-08-07 API~\cite{singh2025openai} with minimal reasoning effort seeting as either ``possible'' or ``impossible'' under the given context.

The following filtering procedure was applied.

\paragraph{Step 1: Gold statute validity verification (Base Confirmation).}

We first verified whether the question could be correctly answered using only the gold statute. Specifically, we checked whether the conclusion derived with the expert answer included (A) was identical to the conclusion derived using only the gold statute (B), i.e., $A = B$. Only samples satisfying this condition were passed to the next stage. This step also served to remove consultation cases written under outdated legal provisions that conflict with current law.

\paragraph{Step 2: Candidate negative extraction (co-citation based).}

Using real precedent data, we extracted statutes that frequently co-occurred with the gold statute but were not themselves the correct statute. Through this process, a total of 731 co-occurring statutes were collected.

\paragraph{Step 3: Conflicting conclusion filtering.}

Each candidate statute was provided to the model to generate the conclusion $B_{conf}$. Only statutes for which the resulting conclusion was clearly different from the gold-statute conclusion ($B_{conf} \neq B$) were retained. This indicates that the candidate statute can lead the model toward a legal interpretation that differs from the correct answer. After this step, 370 statutes remained.

After passing through the entire pipeline from Step 1 to Step 3, a total of 22 consultation cases remained for which valid hard negative options were obtained.

\subsubsection{\tsuff\ Benchmark Question Generation}

Based on the final 22 consultation cases and their corresponding negative statutes, benchmark questions were generated under three different conditions by varying the composition of legal information provided to the model.

\paragraph{Sufficient condition.}

All statutes necessary for the correct answer were provided in the context (already referenced laws). The answer options consisted of three incorrect statutes and ``No additional legal reference.''

\paragraph{Insufficient condition.}

One statute obtained through hard negative mining was provided in the context. The answer options consisted of one correct statute, two incorrect statutes, and ``No additional legal reference.'' In this condition, the model must recognize that the provided statute is insufficient or inappropriate and select the necessary statute.

\paragraph{Partial condition.}

For cases where multiple statutes were required for the correct answer, only part of the gold statutes were provided in the context, while the remaining gold statute was included in the answer options. The model must determine that additional information is required.

For each consultation case, multiple problem variants were generated by combining different negative statutes. In total, 279 benchmark questions were constructed, consisting of 49 Sufficient cases, 123 Insufficient cases, and 107 Partial cases.

\subsubsection{Manual Review and Validation}

All annotations and summaries generated using LLMs were manually reviewed by the researchers. If the legal interpretation was ambiguous or the model output was unstable, the corresponding case was excluded from the dataset.

Finally, the correctness of the answers and the composition of the options for all benchmark questions were confirmed through manual review by the researchers. Through this multi-stage verification process, each benchmark problem was constructed to accurately reflect the intended legal reasoning task.

%% file: Sections/8_appendix_example.tex
\clearpage
\onecolumn

\subsection{Benchmark Examples}

In this section, we present example questions for each task type.
All examples were originally constructed in Korean and are presented here in English translation for readability.

\subsubsection{\ttime\ Example}

\paragraph{Example (translated from Korean)}
\begin{quote}
[Unconstitutionality] For the same case as the following judgment, choose the point in time at which a conclusion similar to the following judgment could be reached.

Case summary: In a case where a doctor, at the request of a woman, performed an abortion on an early-stage fetus by suction curettage, the court found the doctor guilty but suspended the sentence.

Change (timeline premise of constitutional $\rightarrow$ unconstitutional): Due to the Constitutional Court's constitutional nonconformity decision, the criminal-law provisions related to abortion retroactively lost effect, and accordingly a prosecution based on those provisions came to be regarded as not constituting a crime.
Premises:
\begin{itemize}
    \item If the phrase `retrial' does not appear in the options, the trial time refers to the original trial time as it is.
    \item If `retrial' is included in an option, the `trial time' here refers to the trial time at the time of the final and binding judgment.
    \item Let A1 = date of constitutional decision, and A2 = date of unconstitutional decision.
\end{itemize}

Options:
\begin{itemize}
    \item[A.] Assuming a retrial, the time of the act is before A1, and the time of the previous final judgment is between A1 and A2
    \item[B.] Assuming a retrial, both the time of the act and the time of the previous final judgment are between A1 and A2
    \item[C.] Both the time of the act and the time of the trial are before A1
    \item[D.] Both the time of the act and the time of the trial are after A2
\end{itemize}

You must output only one letter among A/B/C/D as the answer. Do not output any other explanation.
\end{quote}

\subsubsection{\tsuff\ Example}

\paragraph{Example (translated from Korean)}
\begin{quote}
Based on the presented `legal consultation question' and the `already referenced law,' choose the legal provision that should be additionally referenced with the highest priority in order to resolve the consultation. Output only one of the options (A/B/C/D).

[Question]

Legal consultation question: After I retired from my company, I did not receive my retirement pay, and recently the company entered bankruptcy proceedings. In this case, I would like to know whether I must receive the money through the bankruptcy procedure or whether I can receive payment at any time as an estate claim.

Already referenced law: Debtor Rehabilitation and Bankruptcy Act, Article 475 (Article 475 (Payment of estate claims): Estate claims are paid at any time without going through bankruptcy procedures.)

[Options]
\begin{itemize}
    \item[A)] Debtor Rehabilitation and Bankruptcy Act, Article 473 (Article 473 (Scope of estate claims))
    \item[B)] Labor Standards Act, Article 36 (Article 36 (Settlement of money and valuables): When a worker dies or retires, the employer shall pay wages, compensation, and all other money and valuables within 14 days from the date the cause for payment arose. However, if there are special circumstances, the due date may be extended by agreement between the parties. <Amended 2020.5.26>)
    \item[C)] Commercial Act, Article 101 (Article 101 (Definition): A commission merchant is a person who, in his or her own name and for the account of another, engages in the business of selling and purchasing goods or securities.)
    \item[D)] No additional legal reference is needed (unnecessary, or none of the other provisions is appropriate to reference)
\end{itemize}

You must output only one letter among A/B/C/D as the answer.

Do not output any other explanation, whitespace, or punctuation mark.
\end{quote}

\subsubsection{\tshift\ Example}

\paragraph{Example (translated from Korean)}
\begin{quote}
Read the summary of the precedent and the description of the change pattern before and after, and answer the question.

[Later Judgment Information]
\begin{itemize}
    \item Facts: The defendant, while in conflict with his wife during marriage and sleeping in separate rooms, assaulted and threatened her with a weapon and the like, making resistance difficult, and then forcibly had sexual intercourse with her; a few days later, he committed a similar offense again. The lower court found that, even though the acts occurred between spouses, a sexual violence crime was established in light of the degree and circumstances of the assault and threat, found him guilty, and also imposed electronic monitoring attachment due to the risk of reoffending.
    \item Conclusion: All appeals are dismissed. / The court found no illegality in the judgment premised on recognizing the crime of rape against the wife and the crime subject to aggravated punishment.
\end{itemize}

[Pattern of Change]
\begin{itemize}
    \item The earlier judgment did not maintain the lower court's judgment as is, taking the position that it was not easy to affirm the establishment of rape for forced sexual intercourse between spouses while the marital relationship was maintained. The later judgment held that even where the marital relationship was maintained, if the assault or threat was enough to make resistance impossible or remarkably difficult, rape was established, and it maintained the guilty judgment.
    \item In a case where, between spouses while the marital relationship continued, the husband forced sexual intercourse accompanied by assault and threat, the establishment of rape was once treated in a limited manner, but later the conclusion changed in the direction of affirming rape and maintaining punishment and ancillary disposition.
\end{itemize}

[Question]

The following is a summary of the later precedent and an explanation of the change in conclusion. What was the most direct cause of the change?

[Options]
\begin{itemize}
    \item[A)] Change in normative premise (change in society's general legal consciousness, common notions, and social/normative foundation)
    \item[B)] Change in normative text (enactment/amendment of relevant legal provisions, including addition/deletion, unconstitutionality/constitutional nonconformity)
    \item[C)] Change in interpretive rule (explicit precedent change)
    \item[D)] Change in factual premise (the normative text and interpretive rule themselves remain the same, but there is creation and change in substantial value, technology/industry environment, etc.)
\end{itemize}

You must output only one letter among A/B/C/D as the answer.
Do not output any other explanation, whitespace, or punctuation mark.
\end{quote}